\documentclass{article}

    \usepackage[preprint]{neurips_2023}

\usepackage[utf8]{inputenc} % allow utf-8 input
\usepackage[T1]{fontenc}    % use 8-bit T1 fonts
\usepackage{hyperref}       % hyperlinks
\usepackage{url}            % simple URL typesetting
\usepackage{booktabs}       % professional-quality tables
\usepackage{amsfonts}       % blackboard math symbols
\usepackage{nicefrac}       % compact symbols for 1/2, etc.
\usepackage{microtype}      % microtypography
\usepackage{xcolor}         % colors

\usepackage{enumitem}
\usepackage{graphicx}
\usepackage{subcaption}
\usepackage{natbib}
\usepackage{algorithm}
\usepackage{algpseudocode}
\usepackage{geometry}

\algtext*{EndWhile}

\usepackage{amsmath,amsfonts,bm}

\def\eqref#1{equation~\ref{#1}}
\def\1{\bm{1}}

\def\vm{{\bm{m}}}

\def\vr{{\bm{r}}}

\def\vx{{\bm{x}}}

\def\vz{{\bm{z}}}

\def\mA{{\bm{A}}}

\def\mI{{\bm{I}}}

\def\mM{{\bm{M}}}

\def\mS{{\bm{S}}}

\DeclareMathAlphabet{\mathsfit}{\encodingdefault}{\sfdefault}{m}{sl}
\SetMathAlphabet{\mathsfit}{bold}{\encodingdefault}{\sfdefault}{bx}{n}

\def\sC{{\mathbb{C}}}

\def\sS{{\mathbb{S}}}

\newcommand{\sancdifi}{\textbf{Sancdifi} }

\hypersetup{
pdftitle={Salient Conditional Diffusion for Defending Against Backdoor Attacks},
pdfsubject={Machine learning},
pdfauthor={Brandon B. May, N. Joseph Tatro, Dylan Walker, Piyush Kumar, Nathan Shnidman},
pdfkeywords={diffusion models, DDPMs, backdoor attacks, adversarial robustness},
}

\title{Salient Conditional Diffusion for Defending Against Backdoor Attacks}

\author{Brandon B. May$^\dagger$ \quad N. Joseph Tatro$^\dagger$\thanks{Corresponding author: joseph.tatro@str.us} \quad Dylan Walker \quad Piyush Kumar \quad Nathan Shnidman   \\
	Vision \& Image Understanding \\
        Systems \& Technology Research \\
	Woburn, MA 01801 \\
}

\begin{document}

\def\thefootnote{$\dagger$}\footnotetext{Equal Contribution}

\maketitle

\begin{abstract}
We propose a novel algorithm, \textbf{Salient Conditional Diffusion} (\textbf{Sancdifi}), a state-of-the-art defense against backdoor attacks. \sancdifi uses a denoising diffusion probabilistic model (DDPM) to degrade an image with noise and then recover said image. Critically, we compute saliency-based masks to condition our diffusion, allowing for stronger diffusion on the most salient pixels. As a result, \sancdifi is highly effective at diffusing out triggers in data poisoned by backdoor attacks. At the same time, it reliably recovers salient features when applied to clean data. This performance is achieved without requiring access to the model parameters of the Trojan network, meaning \sancdifi operates as a black-box defense.
\end{abstract}

\section{Introduction}
With the increasing societal adoption of deep learning models, adversarial robustness, the ability of deep neural networks (DNNs) to withstand adversarial attacks, has quickly become a topic of interest in the machine learning community \citep{madry2017towards}. As the field develops, attention is being given to sophisticated attacks that align more closely to practical use cases. In this work, we focus on defending against backdoor attacks introduced in \textit{BadNet} in \citet{gu2017badnets} as it is particularly challenging to defend against \citep{Li2020BackdoorLA}. In essence, a backdoor attack involves the passing of poisoned data, such as an image containing a visual trigger, to a malicious network that performs adversarially in the presence of said trigger. This creates a striking scenario where an adversary can hack with precision into a seemingly innocuous \textit{Trojan network} that they have released to the public. 

In this work, we:
% \begin{enumerate}[leftmargin=*]
% \itemsep0em 
% \item 
\textbf{(1)} Propose a novel defense against backdoor attacks, \textbf{Sancdifi}, that \textit{purifies} input with a diffusion model (DDPM) conditioned on a mask derived from input dependent saliency maps. 
% \item 
\textbf{(2)} Establish state-of-the-art performance among backdoor defenses. While \sancdifi is a black-box defense, needing no explicit access to the model parameters of the Trojan network, it is competitive with 
% state-of-the-art white-box defenses such as 
fine-pruning \citep{fineprune2018} and Neural Attention Distillation \citep{li2021neural}.  
% \item 
\textbf{(3)} Demonstrate the utility of salient conditioning in our novel algorithm. We experimentally find that less salient parts of an image create a strong prior for the reverse diffusion process of a DDPM. This allows us to more reliably recover clean salient parts of an image.   
% \item Provide visualization of the salient conditional diffusion algorithm and its defensive power against Trojan networks associated with backdoor attacks. 
% \end{enumerate}

Given its straightforwardness and ease-of-use, we are encouraged by the performance of salient conditional diffusion. In this work, we first review related work and provide background. Next, we motivate and formally introduce our algorithm \sancdifi and empirically validate its performance.  

\begin{figure*}[t]
\centering
\includegraphics[width=0.95\textwidth]{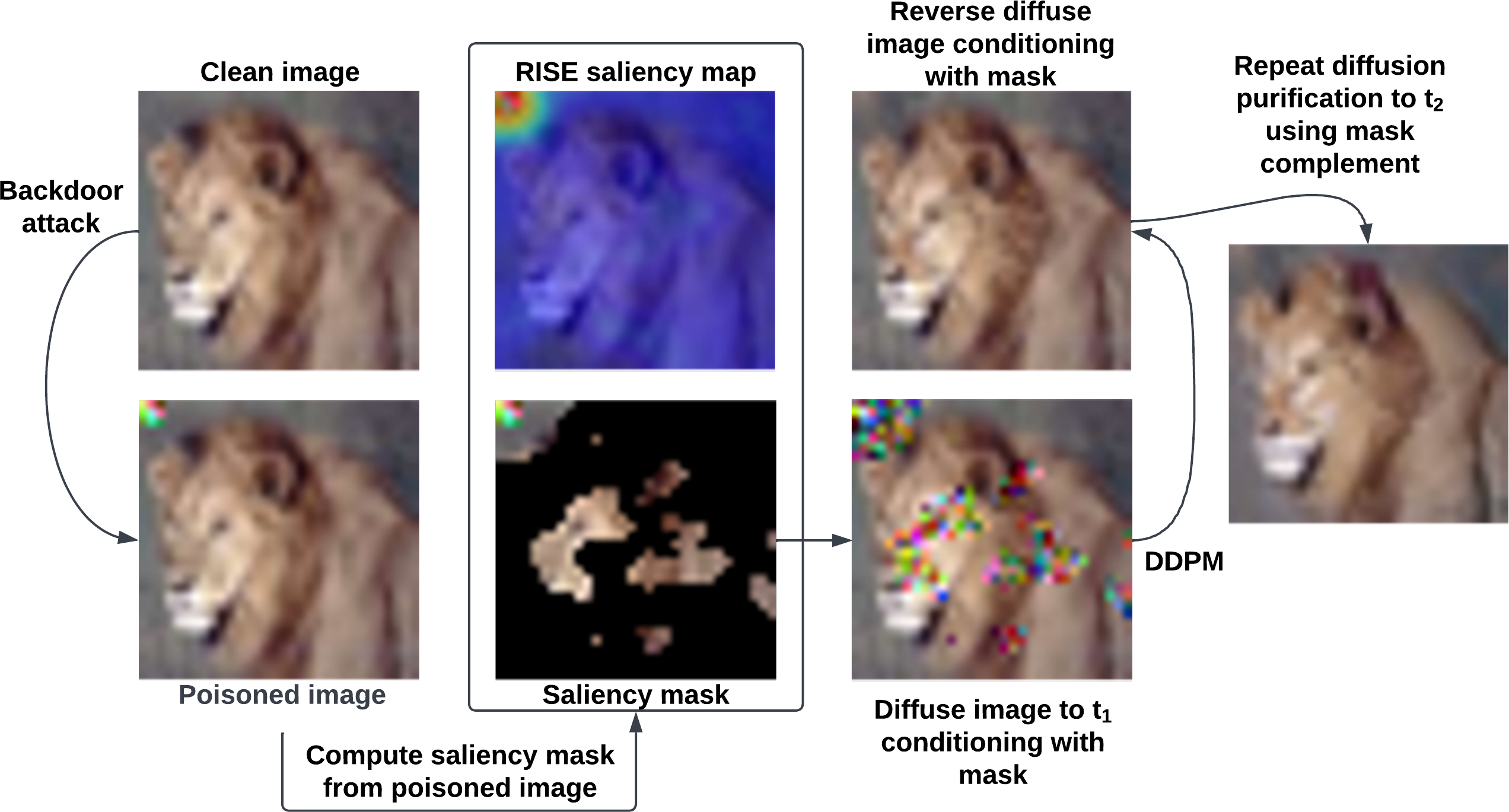}
\caption{An illustration of Salient Conditional Diffusion (\textbf{Sancdifi}). First a trigger is added to a clean image in an attempted BadNet backdoor attack. With this input image, we compute saliency maps via RISE, and use the top-5 class maps to construct the binary visible mask, $\mA$. Notice the trigger is unmasked as it is the most salient part of the top-1 map. We then apply diffusion purification, conditioned with the saliency mask, to the image for 300 time steps. Following this, we reapply diffusion purification using the reverse mask, $\mI - \mA$, with 100 time steps. Notice that \sancdifi is capable of diffusing out the backdoor trigger without largely degrading the entire image.}
\label{fig:sancdifi_main}
\end{figure*}

\subsection{Related Work}
\label{sec:Rel}

% \subsubsection{Backdoor Attacks}
% \label{subsec:backdoor}
\paragraph{Backdoor Attacks} As stated previously, a backdoor attack involves an adversary training a malicious network such as BadNet in \cite{gu2017badnets}. Generally, this \textit{Trojan network} contains a subnetwork that adversarially alters output in the presence of a specific trigger in the input. The detection of these Trojan networks is an ongoing research topic and is a difficult task given the vast permutations of possible subnetworks \citep{wang2020practical}. Recent work has demonstrated that even diffusion models are susceptible to backdoor attacks \citep{chou2022backdoor}. 

Surveyed in works including TrojanZoo \citep{trojanzoo2022} and BackdoorBox \citep{li2023backdoorbox}, there are several types of defenses against backdoor attacks; input reformation, input filtering, model sanitization, and model inspection. 
Our proposed algorithm is an input reformation defense, meaning we prefilter network input while having no knowledge of model weights. Other state-of-the-art input reformation defenses include ShrinkPad (SP) \citep{li2021backdoor_shrinkpad}. Two state-of-the-art methods in backdoor defense, fine-pruning (FP) \citep{fineprune2018}, and Neural Attention Distillation (NAD) \citep{li2021neural}, are model sanitization defenses. 
These two methods are white-box methods that alter the model weights of the Trojan network. We stress that, as a black-box method, \sancdifi is applicable in more general real-world scenarios where fine-pruning is not. Concerning saliency-based methods, Februus \citep{doan2020februus} is a state-of-the-art method that uses a generative adversarial network (GAN) to inpaint an image after applying a GradCAM-derived mask. This use of Grad-CAM for trigger detection can also be found in SentiNet \citep{chou2020senti}. We will show later that \sancdifi successfully defends against more general attacks where Februus fails.     

% \subsubsection{Diffusion Models and Conditioning}
\paragraph{Diffusion Models and Conditioning} Diffusion models such as denoising diffusion probabilistic models (DDPMs) \citep{ho2020denoising} have quickly rivaled generative adversarial networks (GANs) \citep{goodfellow2020generative} in the task of image generation. 
DDPMs act by diffusing input through iteratively adding Gaussian noise and then learning the reverse diffusion process to recover the input image. This reverse diffusion process is able to generate data from noise in a Markov chain-like fashion. 
Research on diffusion models has exploded in the past few years. \citet{kingma2021variational} analyzed the theoretical properties of the variational lower bound of DDPMs. \citet{song2020denoising} introduced an implicit variation, DDIM. Crossing into mainstream awareness, \textit{Stable Diffusion} has powered generative AI apps used by the general public \citep{stablediffusion2022}. 
% Diffusion models find themselves used in various other tasks such as super-resolution \citep{saharia2022image}. 

Of particular interest, \citet{nie2022diffusion} and \citet{wu2022} both proposed diffusion purification, the use of diffusion models for adversarial purification against projected gradient descent (PGD) attacks. A strong inspiration for this work, DiffPure \citep{nie2022diffusion} is able to filter out PGD attacks with a small timescale diffusion and reverse diffusion process governed by a DDPM. We note that a defense appropriate for PGD attacks is not necessarily a valid defense against backdoor attacks \citep{Weng2020OnTT}. Thus, we were inspired to see if diffusion purification could extend to backdoor attacks. In Section \ref{sec:num_exp}, we discuss the need for longer diffusion than in \citet{nie2022diffusion} to sufficiently degrade triggers, while our novel salient conditioning is key to preventing a collapse in clean accuracy.

There has also been work that conditions the diffusion process of DDPMs. \citet{dhariwal2021diffusion} introduced classifier-guided DDPMs. These models condition diffusion with a classifier gradient to steer the generative process to a user-specified class. \citet{voynov2022} introduced a DDPM guided by the gradients of a latent edge predictor to improve text-to-image generation. These works focus on salient features when generating images. Unlike these methods, our work does not require user input, such as class, as a prior. Regarding mask-based conditioning, \citet{aberman2022} introduced a model for reducing saliency within a region of an image determined via user-specified mask. Our algorithm has a different flavor as our mask is determined via saliency map and we do not decimate all salient features within the unmasked region of our images.

\begin{figure}[t]
\centering
\begin{subfigure}[b]{0.32\textwidth}
    \centering
    \includegraphics[width=\textwidth]{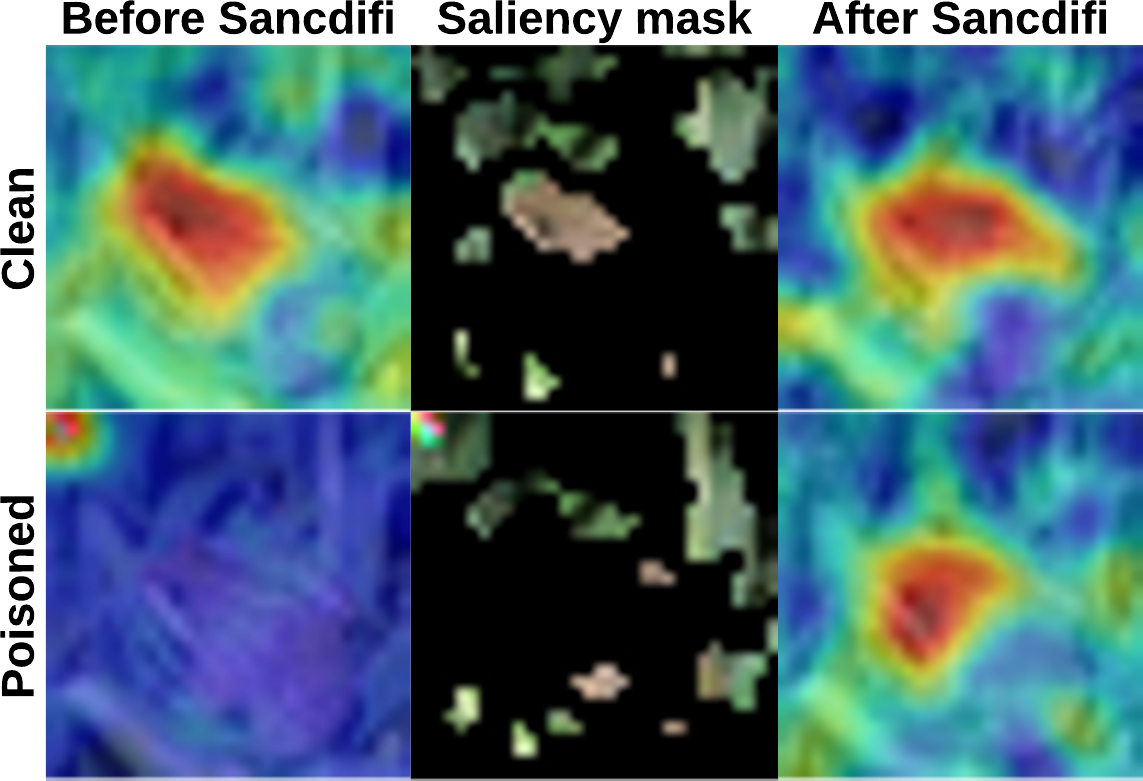}
    \caption{CIFAR-10}
\end{subfigure}
\hfill
\begin{subfigure}[b]{0.32\textwidth}
    \centering
    \includegraphics[width=\textwidth]{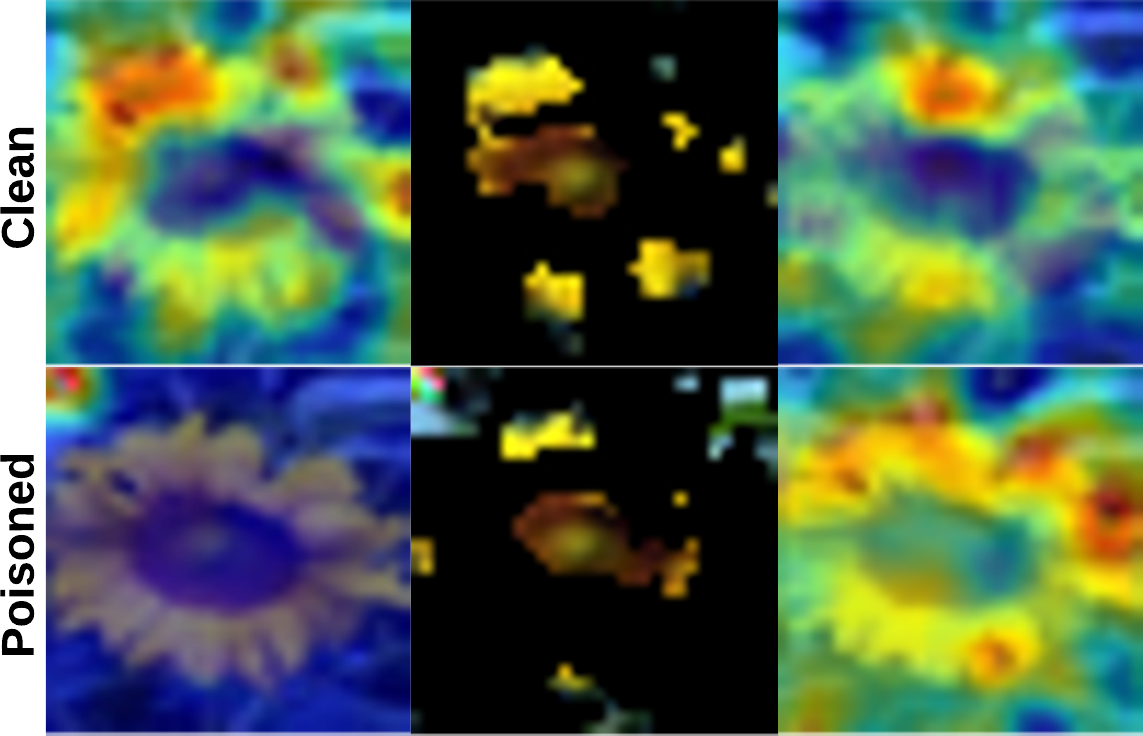}
    \caption{CIFAR-100}
\end{subfigure}
\hfill
\begin{subfigure}[b]{0.32\textwidth}
    \centering
    \includegraphics[width=\textwidth]{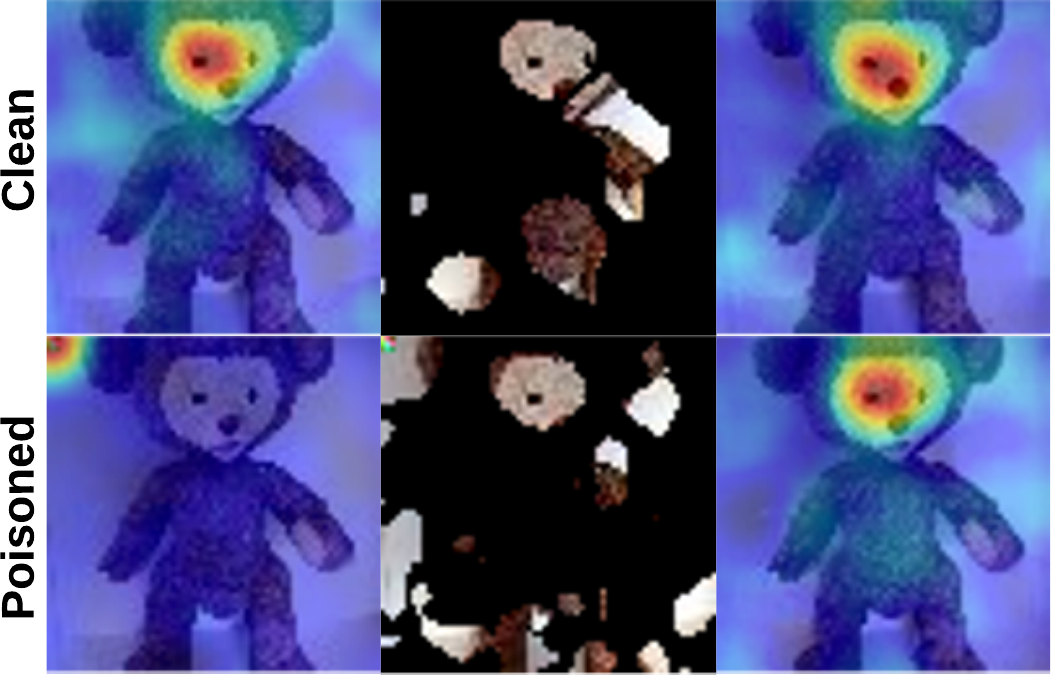}
    \caption{Tiny-ImageNet}
\end{subfigure}
\caption{For CIFAR-10, CIFAR-100, and Tiny-ImageNet respectively, we show the computed saliency mask for ResNet-50. The top and bottom rows respond to clean data and BadNet attacked data. Each column displays; the top class saliency map, the computed saliency mask, and the top class saliency map after applying \textbf{Sancdifi}. Notice we have removed the saliency of the trigger.}
\label{fig:saliency}
\end{figure}

\section{Salient Conditional Diffusion}

To the best of our knowledge, this work is the first to propose the use of diffusion models (DDPMs) as a defense against backdoor attacks. A key contribution of \sancdifi is the use of saliency masks for conditioning diffusion purification. Before explaining our salient conditional diffusion algorithm, we state our backdoor attack model and discuss the motivation of our approach. 

\subsection{Backdoor Attack Model}
In this work, we consider malicious networks that are sensitive to the presence of \textit{triggers} in data. Such a backdoor trigger, a small 3x3 patch, can be seen in Figure \ref{fig:sancdifi_main} which illustrates the entire salient conditional diffusion process. As defined in \citet{trojanzoo2022}, the trigger starts with a pattern $p(\vx)$, that may depend on the data $\vx \in \mathbb{R}^d$, of transparency value $\alpha \in [0,1]$. Additionally, certain elements of the data are masked from the trigger following a mask template, $\vm \in \{0,1\}^d$.  The trigger embedded data, $\vx \oplus \vr$, with trigger $\vr$ is defined as:
\begin{equation} \label{eq:triggered_data}
    \vx \oplus \vr := (1-\vm) \odot ((1- \alpha) \vx + \alpha p(\vx)) + \vm \odot \vx.
\end{equation}
Trojan networks are expected to handle both this poisoned data, $\vx \oplus \vr$, and clean data, $\vx$. The clean data is associated with a label, $y$, while the target label for poisoned data is $t$. Let $f_\theta$ denote our Trojan network with parameters $\theta$, and let $\mathcal{L}$ denote the loss function. Then the objective of the Trojan model is to solve the following optimization problem,
\begin{equation} \label{eq:backdoor_model}
\min_{\vr \in R, \theta} \mathcal{L}(\vx \oplus \vr, t) + \lambda \mathcal{L}(\vx, y).
\end{equation}
Here, $R$ denotes the set of candidate triggers and $\lambda$ is a hyperparameter managing the trade-off between clean accuracy and backdoor attack success.

We use BadNet \citep{gu2017badnets}, TrojanNN \citep{Liu2018TrojaningAO}, and WaNet \citep{nguyen2021wanet} as our attack models. In BadNet, the trigger $\vr$ is fixed, in contrast to other attacks that optimize some combination of the parameters in \eqref{eq:triggered_data}. For instance, TrojanNN optimizes the pixel values of the trigger to maximize certain neuron activations in the Trojan network. While the first two have input-invariant triggers, WaNet is input-adaptive and uses an imperceptible warping-based trigger. All attacks retrain a benign model $f$ where a subset of the data has been poisoned. 
We also consider \textit{Invisible BadNet}, an attack where the trigger has image-wide support but is $L_\infty$-bounded to make it visually imperceptible. 

\begin{figure*}[t]
\centering
\begin{subfigure}[b]{0.33\textwidth}
    \centering
    \includegraphics[width=\textwidth]{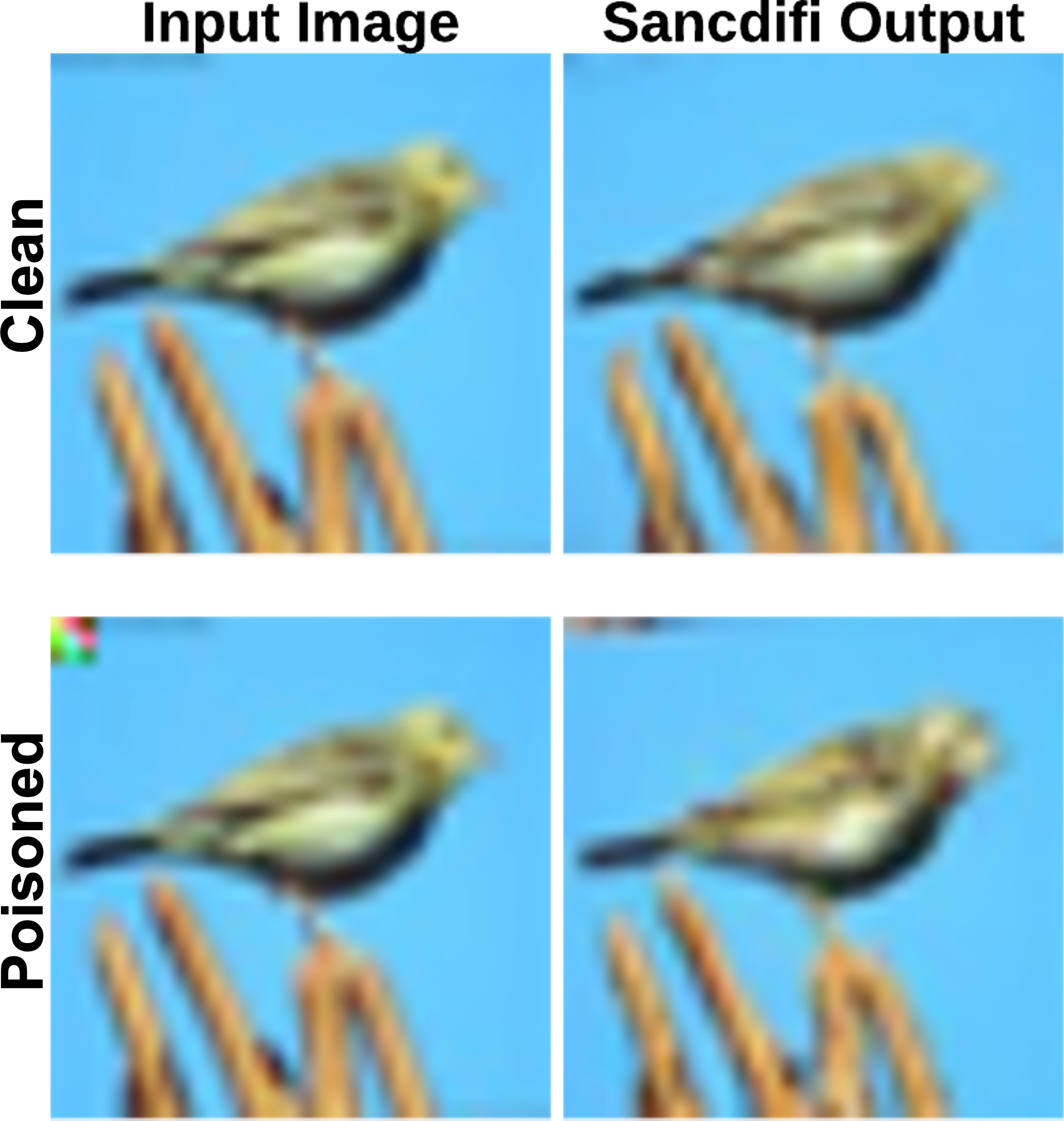}
    \caption{CIFAR-10}
\end{subfigure}
\hfill
\begin{subfigure}[b]{0.315\textwidth}
    \centering
    \includegraphics[width=\textwidth]{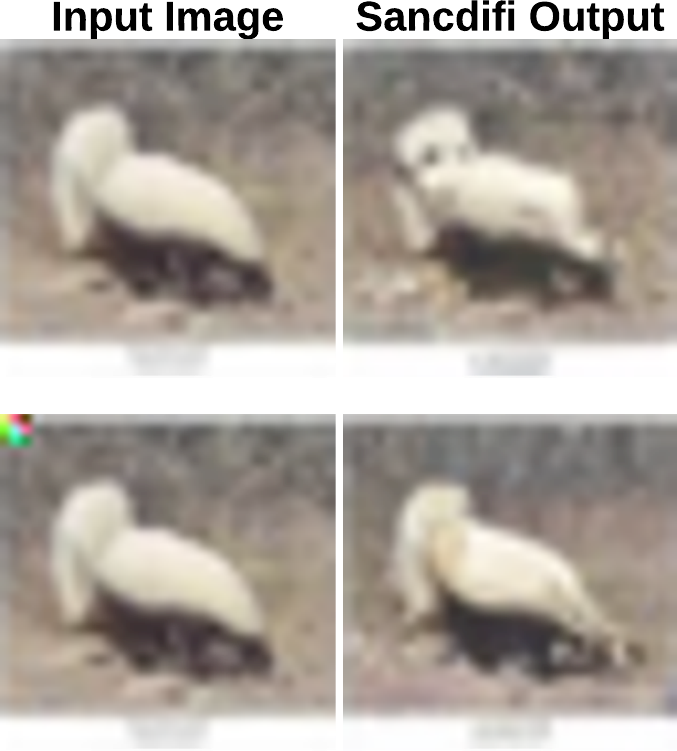}
    \caption{CIFAR-100}
\end{subfigure}
\hfill
\begin{subfigure}[b]{0.315\textwidth}
    \centering
    \includegraphics[width=\textwidth]{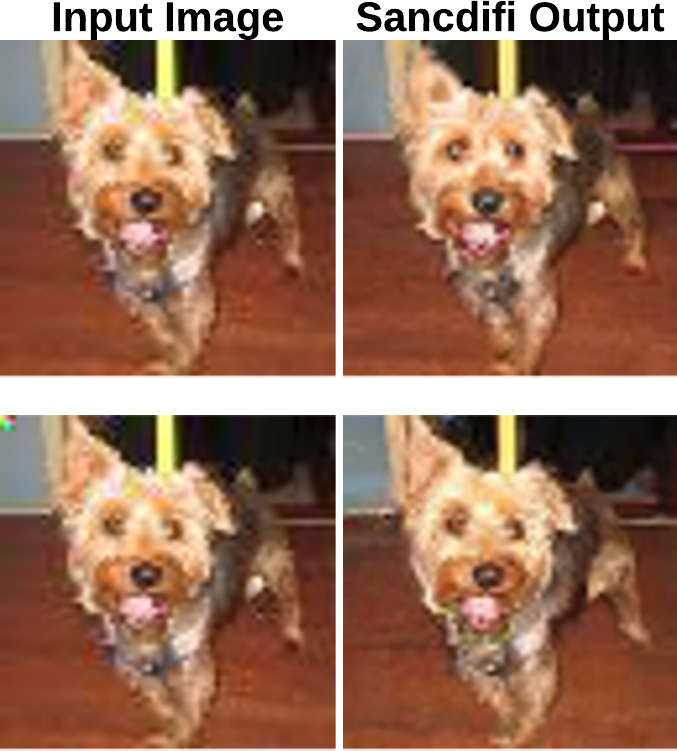}
    \caption{Tiny-ImageNet}
\end{subfigure}
\caption{\sancdifi defending against BadNet attacks on CIFAR-10, CIFAR-100, and Tiny-ImageNet for ResNet-50. The top and bottom rows display it operating on clean and the corresponding BadNet attacked images respectively. Our method removes the trigger from poisoned data while preserving the integrity of clean data.}
\label{fig:badnet_results}
\end{figure*}

\subsection{Motivation}

As visualized in Figure \ref{fig:sancdifi_main}, one of the most common trigger varieties in backdoor attacks are localized triggers. In the figure, the trigger is a 3x3 pixel signature within a 32x32 pixel image. Let $\hat{\vr}:= \vx \oplus \vr - \vx$ denote the visible trigger. Then the support of $\hat{\vr}$ is fairly concentrated. Due to time-frequency duality as described in \citet{chaparro2015frequency}, the frequency spectra of $\hat{\vr}$ can be expected to have wide support. This has been confirmed empirically in \citet{zeng2021}.

It is well-known that (forward) diffusion operates on a function $f$ by exponentially decaying its component frequency modes, with the rate of decay scaling with frequency. It follows diffusion can be used to degrade the high-frequency rich trigger associated with a backdoor attack. Unfortunately, diffusion also degrades the rest of the image. Then it is the reverse diffusion process of a DDPM, trained on clean data, that can recover the diffused image. Since images poisoned with this trigger are assumed not to be in the training data for the DDPM, we expect the trigger to be significantly degraded. It will be this degradation that prevents the activation of the the Trojan network. In order to minimize image degradation from the process, we would like to restrict diffusion to pixels associated with the Trojan trigger or those most likely to be recovered by the DDPM. \sancdifi is motivated by the idea that these pixels correspond with the most salient pixels in an image. 

\begin{algorithm}[t]
\caption{Salient Conditional Diffusion algorithm with image $\vx$, Trojan network $f$, $N$ RISE masks, time steps $\{T_1, T_2\}$, saliency percentile cutoff $d$, and $r$ of top-r performance.}\label{alg:sancdifi}
\begin{algorithmic}
\Require $T_i \geq 0$, $d \in (0,1)$, $r \geq 1$, $i=1$, trained DDPM to parameterize $p(\hat{\vx}_{T-1}| \hat{\vx}_T)$
\State $\sC \gets \text{top-k}(f(\vx), r)$ indices
\State $\sS \gets$ \{RISE($\vx$, $f$, $N$, $c$), \quad $c \in \sC$ \} \Comment{See \citep{petsiuk2018rise} for RISE algorithm}
\State $\mM \gets$ \{$\mS_i \leq percentile(\mS_i, d), \quad \mS_i \in \sS\}$
\State $\mA \gets \prod_i \mM_i, \quad \mM_i \in \mM$
\While{$i \leq 2$} 
\State $\vz \gets $ sample $q(\vx_{T_i}|\vx_0)$ \Comment{defined in \eqref{eq:forward_diff}}
\State $\hat{\vx}_{T_i} \gets \mA \vx_0 + (\mI -\mA) \vz$
\While{$T_i \neq 0$}
\State $\vz \gets$ sample $p(\hat{\vx}_{T_{i}-1}| \hat{\vx}_{T_i})$ \Comment{defined in \eqref{eq:backward_diff}}
\State $\hat{\vx}_{T_i-1} \gets \mA \hat{\vx}_{T_i} + (\mI-\mA) \vz$
\State $T_i \gets T_i - 1$
\EndWhile
\State $\mA \gets \mI - \mA$
\State $i \gets i + 1$
\EndWhile
\end{algorithmic}
\end{algorithm}

\subsection{Methodology}

With the use of DDPMs for defending against backdoor attacks motivated, we describe the \sancdifi algorithm, with a summary being available in Algorithm \ref{alg:sancdifi} and visible in Figure \ref{fig:sancdifi_main}. A core component of our algorithm is the use of saliency to condition diffusion purification. 

A saliency map $\mS_k$ for a given image $\vx$, class $k$, and classifier network $f$ measures the importance of each pixel of $\vx$. This importance is relative to $f's$ determination of the $k$-class probability of $\vx$. Arguably the most well-known saliency map algorithm is the white-box algorithm Grad-CAM, which defines saliency as the gradient of the $k$-class classifier output $\nabla_{\vx} f_k$ \citep{selvaraju2017grad}. We measure saliency using maps generated by the RISE algorithm \citep{petsiuk2018rise}. RISE saliency maps are computed in a black-box fashion, approximating Grad-CAM output, while requiring no knowledge of the model parameters of the network.

Given an input image $\vx$, \sancdifi starts by computing the RISE saliency maps of $\vx$ for the top $r$ classes determined by the Trojan network $f_\theta$. Examples of RISE saliency maps can be seen in Figure \ref{fig:saliency}. The most probable saliency map for clean images highlights meaningful pixels such as the body of a frog, the petals of a flower, or the face of a stuffed animal.
In contrast, the most probable map for BadNet-poisoned images has the strongest response on the trigger. Encouragingly, we will show in Section \ref{sec:results} that the application of our defense to poisoned images produces saliency maps close to their clean counterparts.

From the $k$-class saliency map $\mS_k$, we threshold the top $d$ percentile of values to create a $k$-class saliency mask, $\mM_k$. We desire our algorithm to have robust performance over different validation metrics such as top-5 accuracy. With that in mind, given the set of masks corresponding to the top-$r$ most probable classes $\sS_\mM$, we can define a composite saliency mask $\mA$ as their elementwise product. Concretely,
\begin{equation}
 \label{eq:composite_mask}
 \mA := \prod_{\mM \in \sS_\mM} \mM \quad \text{where} \quad \mM_k := \mS_k \leq percentile(\mS_k, d).   
\end{equation}
We will use $\mA$ to condition our diffusion processes. Intuitively, the composite mask ignores all but the most salient pixels of the most likely classes.  

With the creation of our saliency mask, $\mA$, we discuss the diffusion purification process motivated by \citet{nie2022diffusion}. Our method of diffusion is taken from OpenAI's improved-diffusion DDPM \citep{nichol}. Given input data $\vx$, we begin by diffusing it to time $t$ by sampling from the distribution $q(\vx_t|\vx_0)$ and then applying the mask,
\begin{align}
\label{eq:forward_diff}
q(\vx_t|\vx_0) &:= \mathcal{N}(\vx_t; \sqrt{\hat{\alpha}_t} \vx_0, (1 - \hat{\alpha}_t)\mI) \quad \text{where} \quad \hat{\alpha}_t := \prod_{i=0}^t (1 - \beta_i) \\
\label{eq:forward_output}
    \hat{\vx}_t &= \mA \vx_0 + (1 - \mA) \vz, \quad \vz \sim q(\vx_t|\vx_0).
\end{align}
Here $\beta_t$ denotes the variance schedule of the diffusion process. In practice, we define $\beta_t$ to linearly increase over time. Note that the effect of applying the saliency mask to the output is equivalent to conditioning the diffusion process.

Letting $\hat{\beta}_t$ be $\frac{1-\hat{\alpha}_{t-1}}{1-\hat{\alpha}_t} \beta_t$, the reverse diffusion process involves both the following prior and posterior conditional distributions, $p$ and $q$,:
\begin{align}
    \label{eq:backward_diff}
    p(\vx_{t-1}| \vx_t) = \mathcal{N}(\vx_{t-1}; \mu(\vx_t, t), \hat{\beta}_t \mI), \quad
    &\mu := \frac{1}{\sqrt{1 - \beta_t}}\left(\vx_t - \frac{\beta_t}{1-\hat{\alpha}_t} \mathcal{E}_\theta(\vx_t, t)\right), \\
    \label{eq:backward_posterior}
    q(\vx_{t-1}| \vx_t, \vx_0) = \mathcal{N}(\vx_{t-1}; \hat{\mu}(\vx_t, \vx_0), \hat{\beta}_t \mI), \quad 
    &\hat{\mu} := \frac{\sqrt{\hat{\alpha}_{t-1}} \beta_t}{1 - \hat{\alpha}_t} \vx_0 + \frac{\sqrt{1 - \beta_t} (1 - \hat{\alpha}_{t-1})}{1 - \hat{\alpha}_t} \vx_t. 
\end{align}
To determine $\hat{\vx}_0$, we iteratively sample from the prior distribution $p$. 
In the DDPM framework of \citet{ho2020denoising}, the function $\mathcal{E}_{\theta}$ is parameterized by a neural network. Training a DDPM involves optimizing $\mathcal{E}_{\theta}$ to minimize the sum of KL-divergences of the conditional posteriors from the conditional priors at each time step in the reverse diffusion. 
Conditioning the reverse process is analogous to \eqref{eq:forward_output}, with the critical difference being that masking must take place at each time step. Conditioning diffusion via a mask can be seen in \citet{inpaint2022} for further reference. 

Following this, we reapply the diffusion purification on the resulting image using the complement of our salient mask, $\mI - \mA$, with time $\hat{t}$ where $\hat{t} < t$. This safeguards against attacks with support across the entire image. We diffuse a shorter amount of time as we believe these low-saliency features are less recoverable by the DDPM. We show the importance of this second purification in Section \ref{subsec:perform_other_attacks}. 
% Overall, we believe that our salient conditional diffusion algorithm is fairly intuitive. 

\begin{table*}[t]
\caption{\sancdifi (SD) defense results on BadNet for ResNet-50. Our metrics include clean accuracy reduction (CAR) and attack success rate (ASR) for top-1 and top-5 class performance. CAR represents the drop in accuracy for clean images after applying our defense. We desire \underline{low} values of CAR and ASR. The other algorithms, referenced in Section \ref{sec:num_exp}, are fine-pruning (FP), ShrinkPad (SP), Neural Attention Distillation (NAD), and Februus (FB) respectively. Our algorithm has performance comparable to white-box defenses, FP and NAD, while having our CAR is more consistent across datasets compared to the other black-box defenses.}
\label{tab:badnet_results_resnet}
\begin{center}
\begin{tabular}{llcrrrrrcrrrr} 
\toprule
&&& \multicolumn{10}{c}{\textbf{Backdoor Defenses}} \\
& && \multicolumn{5}{c}{top-1} && \multicolumn{4}{c}{top-5} \\
\cmidrule{4-8} \cmidrule{10-13} \\
\textbf{Dataset}  & \textbf{Metric} && \textbf{SD} & FP & SP & NAD & FB  && \textbf{SD} & FP & SP & NAD \\
% \cmidrule{1-2}  \cmidrule{4-9}
\midrule
CIFAR- & CAR        && 2.0  & -1.0 & 1.0  & 13.0 & 13.0 && 0.0  & 0.0 & 0.0 & 1.0  \\
10 & ASR && 12.0  & 36.0 & 11.0  & 11.0 & 11.0 && 55.0  & 95.0 & 60.0 & 57.0  \\
\midrule
CIFAR- & CAR            && 10.0  & 15.0 & 7.0 & 13.0 & --- && 8.0  & 5.0 & 1.0 & 5.0 \\
100 & ASR && 1.0 & 1.0 & 1.0 & 1.0 & --- && 8.0  & 3.0 & 12.0 & 5.0 \\
\midrule
Tiny & CAR           && 7.0  & 0.0 & 16.0  & 10.0 & --- && 5.0  & 0.0 & 6.0 & 7.0 \\
ImageNet & ASR && 3.0 & 1.0 & 1.0 & 5.0 & --- && 7.0  & 6.0 & 3.0 & 17.0 \\
\bottomrule
\end{tabular}
\end{center}
\end{table*}

\begin{table*}[t]
\caption{\sancdifi (SD) defense top-1 accuracy on BadNet for other networks. Methods and metrics defined are in Table \ref{tab:badnet_results_resnet}. We see that our performance extends to architectures beyond ResNet-50 and is comparable to other defenses. Particularly, we outperform ShrinkPad on these other architectures.}
\label{tab:badnet_results_cifar100}
\begin{center}
\begin{tabular}{llcrrrrcrrr} 
\toprule
&&& \multicolumn{8}{c}{\textbf{Backdoor Defenses}}\\
&&& \multicolumn{4}{c}{CIFAR-100} && \multicolumn{3}{c}{CIFAR-10} \\
\cmidrule{4-7} \cmidrule{9-11}
\textbf{Network}  & \textbf{Metric} & \phantom{a} & \textbf{SD} & FP & SP & NAD && \textbf{SD} & FP & FB  \\
% \cmidrule{1-2}  \cmidrule{4-9}
\midrule
Efficient- & CAR            && 13.0 & 18.0 & 0.0 & 4.0 && 4.0 & -1.0 & 1.0 \\
Net & ASR && 1.0 & 1.0 & 35.0 & 21.0 && 10.0 & 11.0 & 10.0 \\
\cmidrule{1-2}
ViT & CAR            && 17.0 & 1.0 & 3.0 & --- && 5.0 & 0.0 & 1.0  \\
& ASR && 1.0 & 1.0 & 67.0 & --- && 8.0 & 10.0 & 10.0  \\
\bottomrule
\end{tabular}
\end{center}
\end{table*}

\section{Numerical Experiments}
\label{sec:num_exp}

Now that we have discussed the mechanics of our algorithm, we outline the setting of our numerical experiments to validate the performance of \textbf{Sancdifi}. For our experiments, we concern ourselves with the task of image classification in the presence of backdoor attacks. We employ multiple \textbf{datasets}; CIFAR-10, CIFAR-100, \citep{krizhevsky2009learning}, and Tiny-ImageNet \citep{le2015tiny}. Also, we experiment with multiple \textbf{architectures}; ResNet-50 \citep{he2016}, EfficientNet-B7 \citep{tan2019efficientnet}, and a transformer ViT-Base-16 \citep{transformers2020}.
Any metrics reported are for the dataset validation subsets. The networks cover a range of respective qualities; classical and useful, high-performing with low parameterization, and state-of-the-art and complex. Computationally, all experiments were performed on a NVIDIA RTX 2080 Ti GPU.
% and have been seeded for reproducibility. 

Concerning our algorithm, we use pretrained DDPMs from OpenAI's improved-diffusion repository \citep{nichol}. The DDPM we use for a given dataset is trained on that clean dataset. We rescale the timescale of the models from 4000 maximum steps to 1000 steps. 
For the CIFAR datasets, we diffuse out to 300 time steps for the first diffusion purification. For Tiny-ImageNet, we use 450 time steps as we find that it is needed to sufficiently defend against BadNet attacks. For the second diffusion purification step using the complement mask, $\mI - \mA$, we diffuse out to 100 time steps. Our backdoor attacks are generated using the TrojanZoo suite \citep{trojanzoo2022} with their default parameters. In the case of WaNet, we use the BackdoorBox suite \citep{li2023backdoorbox}. 
% For BadNet attacks, we use a preset 3x3 trigger.  

Regarding saliency, we compute RISE maps using 2000 random binary masks. 
% We believe that 2000 RISE masks is sufficient for images with at most 4096 pixels \NS{Why?}. 
For the saliency threshold, we set a value of $95\%$. This cutoff is likely lower than necessary as the 3x3 trigger occupies less than 1\% of an image in our experiments. The composite saliency map is aggregated across the top-5 classes to align us with top-5 metrics.    
For comparison, we evaluate four other defenses against the attacks mentioned in Section \ref{sec:Rel}; fine-pruning (FP) \citep{fineprune2018}, ShrinkPad (SP) \citep{li2021backdoor_shrinkpad}, Neural Attention Distillation (NAD) \citep{li2021neural}, and Februus (FB) \citep{doan2020februus}. Februus results are limited to CIFAR-10 due to availability of inpainting GANs. 
% The first two are state-of-the-art, while the last two are black-box defenses like our own algorithm. 

\subsection{Results}
\label{sec:results}

\begin{table}[t]
\caption{For transferability, \sancdifi results for ResNet-50 using the pretrained ImageNet DDPM. The CIFAR classes are subclasses of ImageNet classes. ImageNet is more complex than the CIFAR as it contains much more data. At the same time, images from ImageNet are in higher resolution. We find that there is some transferability in using the ImageNet DDPM to purify backdoor attacked data from the other related datasets. Both CAR and ASR are higher than in Table \ref{tab:badnet_results_resnet}, but as the BadNet attack is successful 100\% of the time in lieu of any defense, the transfer performance is notable. }
\label{tab:ablation_transfer}
\begin{center}
\begin{tabular}{llcrcrcr} 
\toprule
\textbf{Dataset}  & \textbf{Metric} & \phantom{abc} & baseline top-1 (Table \ref{tab:badnet_results_resnet}) && top-1 && top-5 \\
\midrule
CIFAR-10 & CAR     && 2.0  && 8.0 && 0.0 \\
 & ASR && 12.0 && 17.0 && 59.0 \\
\midrule
CIFAR-100 & CAR         && 10.0  && 24.0 && 14.0  \\
 & ASR && 1.0 && 6.0 && 21.0 \\
\bottomrule
\end{tabular}
\end{center}
\end{table}

\begin{table*}[t]
\caption{\sancdifi (SD) results on other backdoor attacks for ResNet-50. We also include PGD attacks. Clearly, our algorithm can handle other backdoor attacks such as the input-adaptive WaNet as well as the traditional PGD attack. So \sancdifi can be used for both backdoor and adversarial robustness. Our worst scenario is the image-wide Invisible BadNet attack, though we can resolve this by running the second diffusion for longer than 100 steps. For verification, see Table \ref{tab:ablation_ibn_secondtime}.
%NOTE: 30/15 for inviible badnet yields CAR 15/12 and ASR is 3/33
% On regular BadNet, 30/15 maintains ASR and CAR goes to 23/11
}
\label{tab:poisonink_results_resnet}
\begin{center}
\begin{tabular}{llcrrrrcrr} 
\toprule
&&& \multicolumn{7}{c}{\textbf{Backdoor Defenses}} \\
&&& \multicolumn{4}{c}{CIFAR-100} && \multicolumn{2}{c}{CIFAR-10} \\
\cmidrule{4-7} \cmidrule{9-10} \\
\textbf{Attack}  & \textbf{Metric} & \phantom{abc} & \textbf{SD} & FP & SP & NAD && \textbf{SD} & FB \\
\midrule
WaNet & CAR && 14.0 & 17.0 & 20.0 & 0.0 && 5.0 & 36.0 \\
& ASR && 2.0 & 18.0 & 38.0 & 97.0 && 1.0 & 74.0 \\
\midrule
Invisible & CAR        && 5.0 & -3.0 & 1.0 & 3.0 && 3.0  & 2.0   \\
BadNet & ASR && 20.0 & 1.0 & 19.0  & 13.0  && 11.0  & 88.0  \\
\midrule
TrojanNN & CAR            && 9.0 & 4.0 & 8.0  & 2.0  && 5.0  & 8.0   \\
 & ASR && 1.0  & 2.0 & 0.0  & 0.0  && 35.0   & 100.0  \\
\midrule
PGD & CAR            && 18.0  & --- & 3.0  & 8.0 && 0.0  & 7.0      \\
 & ASR && 0.0 & --- & 4.0  & 95.0  && 10.0  & 88.0   \\
\bottomrule
\end{tabular}
\end{center}
\end{table*}

% \subsubsection{Performance on BadNet Attack}
\paragraph{Performance on BadNet Attack}
We first discuss the results of our \sancdifi algorithm in defending against a BadNet attack on ResNet-50. Table \ref{tab:badnet_results_resnet} displays performance in terms of clean accuracy reduction (CAR) and attack success rate (ASR). While we focus mainly on top-1 classification accuracy, we also include top-5 classification results. To be clear, CAR denotes the reduction in accuracy on clean images after applying the defense algorithm. Intuitively, we desire CAR and ASR to be as low as possible. \sancdifi performs comparably to these state-of-the-art approaches. In this case, the winner among these methods is largely a question of the tradeoffs between CAR and ASR as well as top-1 and top-5 performance. The \sancdifi defense associated with these results is visible in Figure \ref{fig:badnet_results}. In the figure, the BadNet trigger has clearly been diffused after purification. In contrast, the other salient regions of the images not covered by salient mask $\mA$ have been reliably recovered by the DDPM.  

To further validate our performance, we repeat the previous experiment for our other architectures on CIFAR-100. We include a CIFAR-10 results for comparison with Februus . Summarized in Table \ref{tab:badnet_results_cifar100}, the behavior of our algorithm is similar to Table \ref{tab:badnet_results_resnet}. Notably, the performance of ShrinkPad does not generalize to other architectures. This shows that the performance of \sancdifi generalizes to various classes of neural networks. 

Regarding the dataset-specific DDPM, Table \ref{tab:ablation_transfer} demonstrates that we still achieve decent performance when using the ImageNet DDPM across datasets. This suggests that a more general domain DDPM can be used in defending related, more specific data. This is helpful as in practice, we may wish to forgo training a DDPM from scratch if a pretrained one for similar data exists.    

\begin{figure}[t]
\centering
\includegraphics[width=0.75\textwidth]{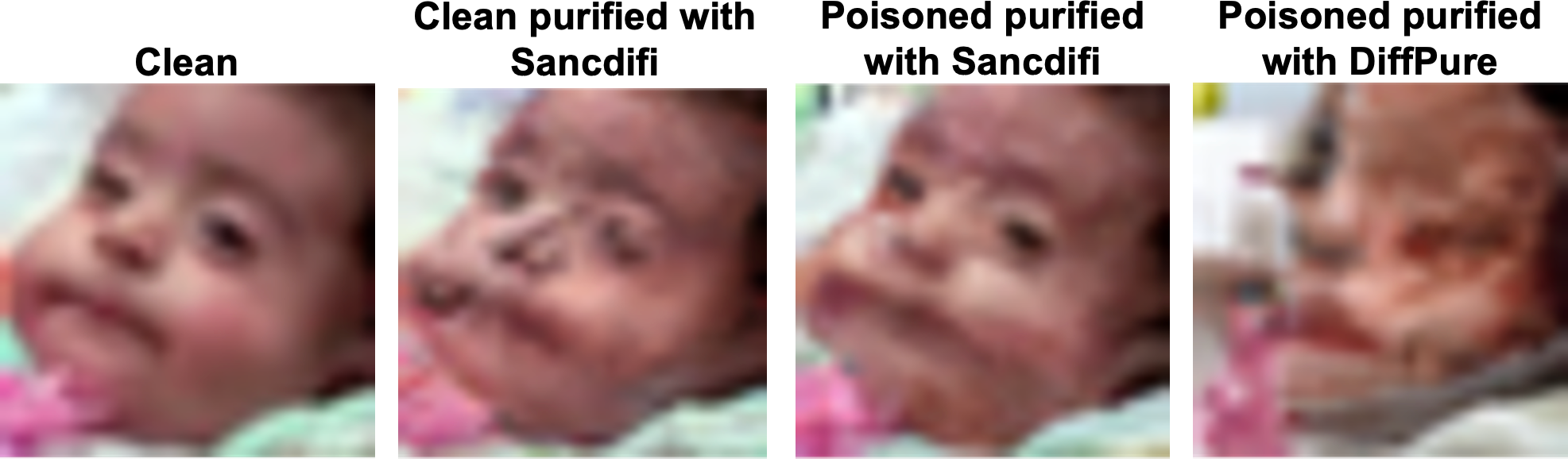}
\caption{Comparison of diffusion purification (\textbf{DiffPure}) to \textbf{Sancdifi}, which includes salient conditioning. This example is with CIFAR-100 and ViT. We display the clean image, the clean image purified with \textbf{Sancdifi}, the BadNet attacked image purified with \textbf{Sancdifi}, and the BadNet attacked image purified with DiffPure. Without salient conditioning, the face is destroyed.}
\label{fig:saliency_impact_diffusion}
\end{figure}

% \subsubsection{Performance against Other Attacks}

\paragraph{Performance against Other Attacks}
\label{subsec:perform_other_attacks}
We have established that \sancdifi achieves competitive performance on BadNet backdoor attacks. For thoroughness, we consider other backdoor attacks; the input-adaptive WaNet, Invisible BadNet, and TrojanNN. We also include the traditional PGD adversarial attack. Table \ref{tab:poisonink_results_resnet} contains our results on defending against these attacks. Our algorithm performs well across the various attacks. We can rectify our weakest performance, which is on Invisible BadNet, by increasing the complement diffusion purification steps past 100 iterations. Table \ref{tab:ablation_ibn_secondtime} in the appendix verifies this is possible without notably harming our effectiveness against regular BadNet attacks.
It is apparent that our defense can prevent PGD attacks better than adversarial retraining of \cite{madry2017towards} which has CAR/ASR of 16.0\%/8.0\% for CIFAR-100. This is important as our algorithm is able to avoid the trade-off between adversarial robustness and backdoor robustness which has been suggested in literature \citep{Weng2020OnTT}. Critically, Februus does not adequately defend against PGD attacks as its inpainting procedure only alters a small portion of the image, in contrast to \sancdifi which diffuses the entire image to some extent. Furthermore, inpainting the entire image is not possible. 
% Similarly, WaNet is known to better preserve the original saliency map during poisoning instead of concentrating saliency in a handful of pixels as in BadNet attacks. Thus, the performance of Februus suffers as larger inpainting area is required. 
With this, \sancdifi has an advantage over the other input-reformation defenses in terms of architecture generalization and adversarial robustness. 

\begin{table*}[t]
\caption{Diffusion results \textbf{without} salient conditioning for ResNet-50. This reduces to the DiffPure algorithm \citep{nie2022diffusion}. Diffusion times are denoted relative to the maximum 1000 time steps. As diffusion time increases, ASR decreases at the cost of increased CAR. At less than 30\% diffusion, ASR can become too high as in the case of Tiny-ImageNet. Yet the high diffusion leads to worse CAR. Notice that in the case of CIFAR-100, CAR is much higher at 30\% than our algorithm (SD) in Table \ref{tab:badnet_results_resnet}. Thus, saliency masking is needed.}
\label{tab:ablation_no_saliency}
\begin{center}
\begin{tabular}{llcrcrcrcrcrcr} 
\toprule
&&& \multicolumn{11}{c}{\textbf{Diffusion Times}}\\
&&& \multicolumn{5}{c}{top-1} & \phantom{ab} & \multicolumn{5}{c}{top-5} \\
 \cmidrule{4-8} \cmidrule{10-14}
\textbf{Dataset}  & \textbf{Metric} & \phantom{ab} & 10\% & \phantom{a} & 20\% & \phantom{a} & 30\% && 10\% & \phantom{a} & 20\% & \phantom{a} & 30\% \\
% \cmidrule{1-2}  \cmidrule{4-9}
\midrule
CIFAR-10 & CAR &  & 5.0 && 5.0 && 9.0 && 1.0 && 1.0 && 2.0 \\
& ASR && 90.0 && 14.0 && 11.0 && 100.0 && 63.0 && 61.0 \\
\midrule
CIFAR-100 & CAR            && 13.0 && 31.0 && 47.0 && 2.0 && 15.0 && 31.0 \\
& ASR && 42.0 && 1.0 && 0.0 && 82.0 && 3.0 && 3.0 \\
\midrule
Tiny & CAR            && 2.0 && 2.0 && 13.0 && 0.0 && 3.0 && 6.0 \\
ImageNet & ASR && 99.0 && 47.0 && 9.0 && 99.0 && 53.0 && 15.0 \\
\bottomrule
\end{tabular}
\end{center}
\end{table*}

\begin{table}[t]
\caption{\sancdifi results for CIFAR-100 \textbf{without} second diffusion purification using the complement mask, $\mI - \mA$. We consider attack scenarios from Table \ref{tab:poisonink_results_resnet}. Invisible BadNet suffers most without the complement diffusion purification, with PGD also increasing in ASR.}
\label{tab:noseconddiff_results_resnet}
\begin{center}
\begin{tabular}{llcrcrcr} 
\toprule
\textbf{Attack}  & \textbf{Metric} & \phantom{abc} & baseline top-1 && top-1 && top-5 \\
\midrule
Invisible & CAR        && 5.0 && 1.0 && 6.0 \\
BadNet & ASR && 20.0 && 84.0 && 99.0 \\
\midrule
TrojanNN & CAR            && 9.0 && 4.0 && 5.0  \\
 & ASR && 1.0 && 1.0 && 4.0 \\
\midrule
PGD & CAR            && 18.0 && 13.0 && 6.0  \\
 & ASR && 0.0 && 8.0 && 33.0 \\
\bottomrule
\end{tabular}
\end{center}
\end{table}

% \subsubsection{Impact of Saliency Masks}
% \label{sec:res_sal_masks}
\paragraph{Impact of Saliency Masks}
% It is reasonable to consider the impact of salient conditioning on our algorithm. 
One might assume that vanilla diffusion purification a la DiffPure \citep{nie2022diffusion} is sufficient against backdoor attacks. Table \ref{tab:ablation_no_saliency} provides results on ResNet-50 where we have performed no salient thresholding and omit the second diffusion purification step. Strikingly, CAR is much worse without salient masking. Notably, DiffPure at 30\% has the worst CAR across all defenses for the CIFAR-100 dataset. 
% We attribute this massive hit to the CAR metric to CIFAR-100 having more classes than CIFAR-10, while being lower resolution than Tiny-ImageNet. 
Additionally, Table \ref{tab:ablation_no_saliency} varies the choice of diffusion time; 10\%, 20\%, and 30\% of maximum time. Lower diffusion times reduce CAR but at the expense of an increased ASR. Thus, salient conditioning is a critical part of our algorithm for succesfully defending against backdoor attacks. A comparison of the output with and without salient conditioning is visible in Figure \ref{fig:saliency_impact_diffusion}. We can see that while it is not the most salient, the masked part of the image offers a strong prior for the DDPM. This prior allows us to more reliably recover the unmasked part of the image excluding the backdoor trigger.  

We also verify the importance of the second diffusion purification step with the complement mask in Table \ref{tab:noseconddiff_results_resnet}. We see that compared to Table \ref{tab:poisonink_results_resnet}, ASR is much higher in Invisible BadNet as well as PGD. Figure \ref{fig:saliency_invisiblebadnet} in the appendix displays how the saliency mask evolves throughout applying \sancdifi to an image that has suffered an \textit{Invisible BadNet} attack. Notice the second diffusion purification shifts focus back towards the left side of the house and the right side of its roof.
 
\section{Conclusion}
We have presented salient conditional diffusion, \textbf{Sancdifi}, a state-of-the-art defense against backdoor attacks. Our algorithm is intuitive with wide generalization over various datasets and network architectures. It is a black-box defense, requiring no knowledge of the parameters of a trojan model, with the performance of state-of-the-art defenses like fine-pruning. We have confirmed its performance against both the classic BadNet attack, imperceptible and input-adaptive WaNet, and TrojanNN attack. Additionally, we find that our algorithm can be used as a preprocessing step to improve the adversarial robustness of a system. Thus we avoid sacrificing adversarial robustness in the pursuit of backdoor robustness. 

Salient conditioning has played a major role in allowing us to diffuse out backdoor triggers without massive degradation to other parts of an image. We believe conditional diffusion will play a strong role in the future in defending against backdoor attacks. 
% In future work, we plan to explore the design of other masks and are interested in extending this work to other models like latent diffusion models (LDMs). 

\begin{ack}
This work was supported by the DARPA AIE program, Geometries of Learning (HR00112290078). 
\end{ack}

\clearpage
% \section*{References}
\bibliography{main}

\clearpage

\appendix

\section{Appendix}

\begin{figure}[h]
\centering
\includegraphics[width=0.7\textwidth]{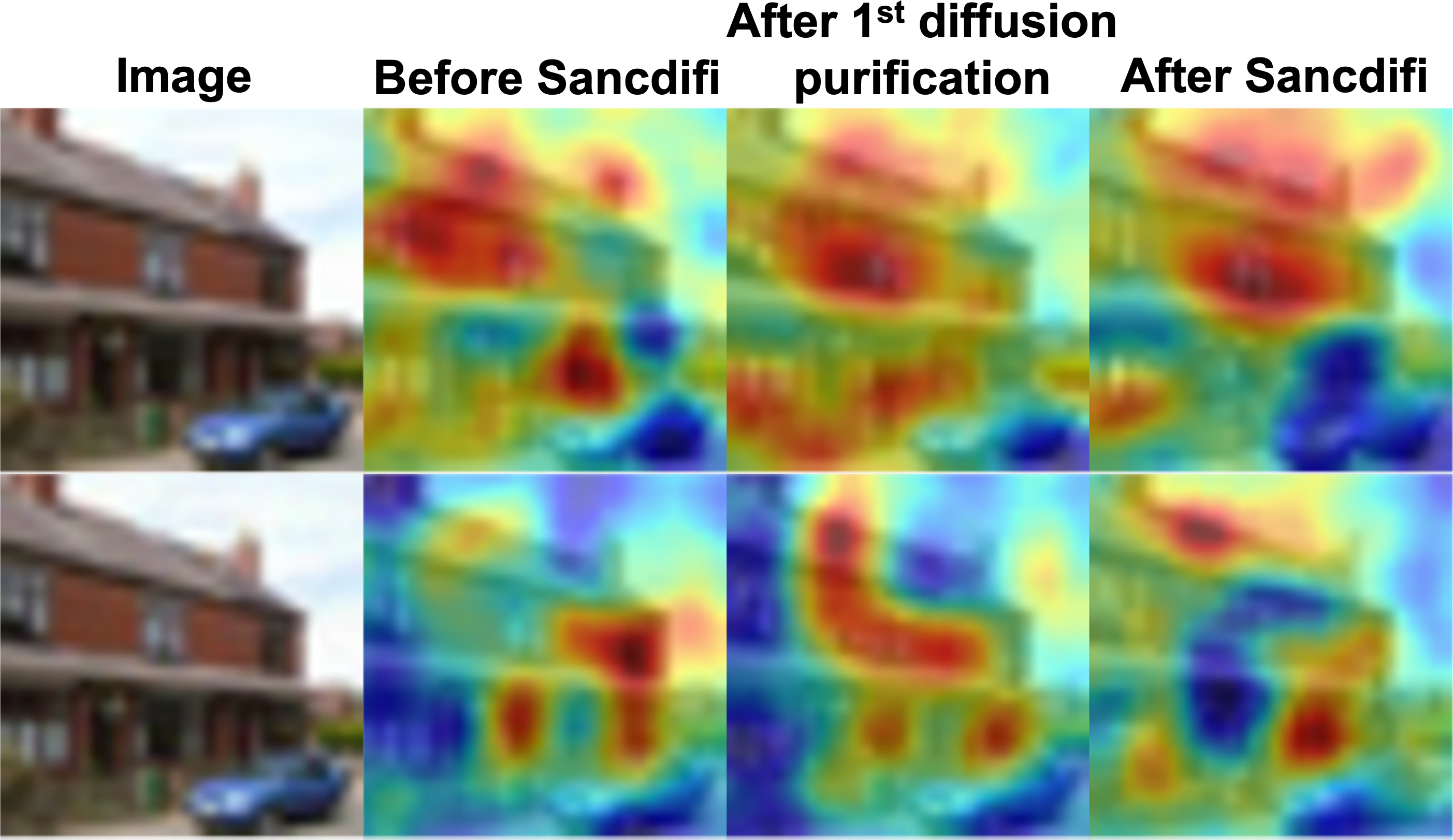}
\caption{Saliency maps displayed for an \textbf{Invisible BadNet} attack on CIFAR100 for ResNet-50. The first and second rows correspond to the clean and attacked images. As the Invisible BadNet attack covers has support of the entire image, the first diffusion purification step does not sufficiently bring the saliency map inline with that of the clean image. We accomplish this with a second diffusion purification at smaller time using the complement mask.}
\label{fig:saliency_invisiblebadnet}
\end{figure}

\begin{table}[h]
\caption{\sancdifi defense results on CIFAR-100 and ResNet-50 with different time steps used for the second diffusion purification. We find that a larger second timescale assists our algorithm in responding to Invisible BadNet attacks. At the same time, the increase in CAR on traditional BadNet attacks is still in line with other methods. 
}
\label{tab:ablation_ibn_secondtime}
\begin{center}
\begin{tabular}{llcrcrccrcr} 
\toprule
& & & \multicolumn{3}{c}{Invisible BadNet} && \multicolumn{3}{c}{Traditional BadNet}  \\
\cmidrule{4-6}  \cmidrule{8-10}
\textbf{Diffusion times}  & \textbf{Metric} & \phantom{abc} & top-1 && top-5 & \phantom{ab} & top-1 && top-5 \\
\midrule
300/100 steps & CAR        && 5.0 && 10.0 && 18.0 && 11.0 \\
 & ASR && 20.0 && 72.0 && 0.0 && 7.0 \\
\midrule
300/150 steps & CAR            && 15.0 && 12.0 && 23.0 && 11.0  \\
 & ASR && 3.0 && 33.0 && 0.0 && 7.0 \\
\bottomrule
\end{tabular}
\end{center}
\end{table}

%%%%%%%%%%%%%%%%%%%%%%%%%%%%%%%%%%%%%%%%%%%%%%%%%%%%%%%%%%%%

\end{document}